\documentclass[runningheads]{llncs}

\usepackage{amsmath,amsfonts,bm}
\usepackage{mathtools}
\usepackage{graphicx}
\usepackage{float}
\usepackage{hyperref} 
\usepackage{cuted}
\usepackage{lipsum, color}
\usepackage{latexsym,amssymb}
\usepackage{multirow}
\usepackage{url}

\usepackage{amsmath, amsthm,epsfig,bm,booktabs,amssymb}
\usepackage[ruled,vlined]{algorithm2e}
\usepackage{amsmath, amsfonts}
\usepackage{amsmath,latexsym,amssymb,epsf,amsfonts,amsthm}
\usepackage{epsf}
\usepackage{color}
\usepackage{mathrsfs}
\usepackage{epstopdf}
\usepackage{threeparttable}
\usepackage{graphicx}
\usepackage{tabularx}
\usepackage[all]{xy}

\newcommand{\no}{\nonumber}

\newcommand{\be}{\begin{equation}}
\newcommand{\ee}{\end{equation}}
\newcommand{\ba}{\begin{eqnarray}}
\newcommand{\ea}{\end{eqnarray}}
\newcommand{\bi}{\begin{itemize}}
\newcommand{\ei}{\end{itemize}}

\newcommand{\comments}[1]{}

\begin{document}
\title{Learning Sparse Neural Networks via $\ell_0$ and T$\ell_1$\\ by a Relaxed Variable Splitting Method \\with Application to Multi-scale Curve Classification}
\titlerunning{Learning Sparse Neural Networks}

\author{Fanghui Xue \and
Jack Xin}

\institute{Department of Mathematics, UC Irvine, Irvine, CA 92697, U.S.A.\\ 
\email{\{fanghuix,jack.xin\}@uci.edu} }

%\thanks{Department of Mathematics, UC %Irvine, Irvine, CA 92097.
% Email: fanghuix@uci.edu.}\;  and 
%\hspace{.1 in} \\
%\date{December 14, 2017.}
%\date{}
\maketitle
%\thispagestyle{empty}
%\newpage 
%\endcsname
\begin{abstract}
We study sparsification of convolutional neural networks \\ (CNN) by a relaxed variable splitting method of $\ell_0$ and transformed-$\ell_1$ (T$\ell_1$) penalties, with application to complex curves such as texts written in different fonts, and words written with trembling hands simulating those of Parkinson's disease patients. The CNN contains 3 convolutional layers, each followed by a maximum pooling, and finally a fully connected layer which contains the largest number of network weights.  With $\ell_0$ penalty, we achieved over 99 \% test accuracy in distinguishing shaky vs. regular fonts or hand writings with above 86 \% of the weights in the fully connected layer being zero. Comparable sparsity and test accuracy are also reached with a proper choice of T$\ell_1$ penalty. 

\end{abstract}
\bigskip

%\hspace{.12 in} {\bf AMS subject classifications:} 92B20, 65K10, 90C26.

%\newpage 

%\setcounter{page}{1}
\section{Introduction}
\setcounter{equation}{0}
Sparsification of neural networks is one of the effective complexity reduction methods to improve efficiency and  generalizability \cite{Welling,DX_18}.  
In this paper, we  sparsify convolutional 
neural networks (CNN) for classifying curves with multi-scale structures. Such curves arise in hand writings of people with neurological disorders e.g. Parkinson disease (PD) patients, and in neuropsychological exams.  Distinguishing hand writings of normal and PD subjects computationally will greatly help diagnosis and reduce physicians' workload in   evaluations.
\medskip

People with PD tend to lose control of their hands, and their writing
or drawing shows oscillatory behavior 
as shown in 
Fig. \ref{PD}, a century old image available online. 
Such oscillatory  features can be learned during CNN training. 
Since we do not have large amount of PD hand writings, we shall generate on the computer a large number of oscillatory shapes that mimic shaky writings of PD subjects. Indeed, we found that CNN is quite successful for this task and can reach accuracy as high as 99 \% on our synthetic data set with three 
convolution layers and one fully connected layer as shown in Fig. \ref{cnn}. However, we also found that there is a lot of redundancy in the weights of the trained CNNs, especially in the fully connected layer where we aim to significantly sparsify the network weights with minimal loss of accuracy. 
\medskip

Since the natural sparsity promoting penalty $\ell_0$ is discontinuous, we shall adopt the relaxed variable splitting method (RVSM, \cite{DX_18}) for network sparsification. Even though Lipschitz continuous penalties such as $\ell_1$ and 
transformed-$\ell_1$ \cite{DMD,ZX} are almost everywhere differentiable, the splitting approach \cite{DX_18} is more effective for enforcing sparsity than directly placing a penalty function inside the stochastic gradient descent (SGD) algorithm. The RVSM is also much simpler than the statistical $\ell_0$ regularization approach in \cite{Welling}. A systematic comparison with \cite{Welling} will be conducted elsewhere. 
\medskip

The rest of the paper is organized as follows. In section 2, we review RVSM for $\ell_0$, transformed-$\ell_1$, and $\ell_1$ penalties and present a convergence theorem.  A new critical point condition is introduced for the limit. We apply RVSM to CNNs for multi-scale curve classification. 
In section 3, we describe our data set, CNN architecture and training, the CNN performance in terms of network accuracy and sparsity. We compare weight distributions of sparse and non-sparse networks. Concluding remarks are in sections 4. 

\section{Sparse Neural Network Training Algorithm}
When training neural networks, one minimizes a penalized objective function of the form:
  \begin{align*}
    l(w) := f(w) + \lambda \, P(w),
  \end{align*}
where $f(w)$ is a standard loss function in neural network models such as cross entropy \cite{YD_15}, and $P(w)$ is a penalty function. In SGD, the expected loss $f$ is replaced by an empirical loss over batches of training samples \cite{YD_15}. In this section, we shall consider the expected loss function $f$ which has better regularity than the empirical loss functions \cite{Yin18}, and is  more conducive to  analysis. In the actual training, SGD and the sample averaged empirical loss function will be implemented. The standard penalty is $\ell_2$ norm, also known as weight decay. However, $\ell_2$ penalty cannot reduce the number of redundant parameters, resulting in a network with on the order of millions of nonzero weights. Thus we turn to $\ell_0$ penalty, which produces zero weights during training \cite{Welling}, however leads to a non-convex discontinuous optimization problem. In \cite{Welling}, a statistical approach is proposed to regularized $\ell_0$. In this paper, 
we utilize the Relaxed Variable Splitting Method (RSVM) studied in \cite{DX_18} for a neural network regression problem. RSVM is much simpler to state and 
implement than \cite{Welling}. To this end, 
let us consider the following objective function for parameter $\beta > 0$:
  \begin{align*}
    \mathcal{L}_{\beta}(u,w) = f(w) + \, \lambda \, P(u) + \frac{\beta}{2}\lVert w-u \rVert^2_2 .
  \end{align*}
Let $\eta$ be the learning rate. We minimize 
$\mathcal{L}_{\beta}(u,w)$ with the RVSM algorithm below where the $u$ step is thresholding and 
the $w$ step is gradient descent followed by a normalization:

    \begin{algorithm}[H]
    \SetAlgoLined
    
       Initialize $u^0$, $w^0$ randomly. \\
     \While{not converged}{    $u^{t+1} \gets \mathop{\arg\min}_{u} \mathcal{L}_{\beta}(u,w^t)$\\
        $\hat{w}^{t+1} \gets w^t - \eta \nabla f(w^t) - \eta \beta (w^t - u^{t+1})$ \\
        $w^{t+1} \gets \frac{\hat{w}^{t+1}}{\lVert\hat{w}^{t+1}\rVert}$
     }
     \caption{RVSM}
    \end{algorithm}

The main theorem of \cite{DX_18} guarantees the convergence of RVSM algorithm under some conditions on the parameters $(\lambda,\beta,\eta)$ and initial weights in case of one convolution layer network and Gaussian input data. The latter conditions are used to prove that the loss function $f$ has Lipschitz gradient away from the origin. Assuming that the Lipschitz gradient condition holds for $f$, we adapt the 
main result of \cite{DX_18} into: 
    \begin{theorem}
    Suppose that $f$ is bounded from below, and satisfies the Lipschitz gradient inequalities: $\|\nabla f(x)-\nabla f(y)\|\leq L_1\, \|x-y\|$, and $|f(x) - f(y) - \langle \nabla f(x), x-y \rangle | \leq L_2\, \|x - y\|^2$, $\forall (x,y)$ with $\|x\|\geq \delta_0$, $\|y\|\geq \delta_0$ for some positive constants $\delta_0$, $L_1$, 
    and $L_2$. Then there exists a positive constant $\eta_0 = \eta_0(\delta_0,L_1,L_2,\beta) \in (0,1)$ so that if $\eta < \eta_0$,
    %be small enough so that $\lVert \eta \nbala f(w^t)+\eta \, \beta (w^t-u^{t+1})\rVert \leq \frac{1}{2}$, for all $t$. 
    %Suppose also that the angle $\theta(w^0,w^*)$ between the initialized weights and the true weights satisfies $\theta(w^0,w^*) \leq \pi -\delta_1$, for some $\delta_1 > 0$. 
    %with $\beta \leq \frac{\delta \sin \delta}{k \pi}$ and $\frac{\lambda}{\beta} < \frac{1}{\sqrt{d}}$. 
    %Then 
    the Lagrangian function  $\mathcal{L}_{\beta}(u^t,w^t)$ is descending and converging in $t$, with $(u^t,w^t)$ of RVSM algorithm satisfying $\|(u^{t+1},w^{t+1}) - (u^t,w^t)\| \to 0$ as $t \to +\infty$, and subsequentially approaching a limit point $(\bar{u}, \bar{w})$.
    %if additionally $\eta \leq \frac{1}{\beta + \beta_1}$, $\beta_1$ being a constant depending on $L_1$ and $L_2$.
    \end{theorem}
\medskip

For the $\ell_0$ penalty, our objective function (the Lagrangian) becomes
  \begin{align*}
    \mathcal{L}_{\beta}(u,w) = f(w) + \lambda \lVert u \rVert_0 + \frac{\beta}{2}\lVert w-u \rVert^2_2 .
  \end{align*}
In this case, we simply obtain 
   \begin{align*}
    u^{t+1} = \mathop{\arg\min}_{u}\mathcal{L}_{\beta}(u,w^t) = H_{\lambda / \beta}(w^t),
  \end{align*}
where $H_{\gamma}$ is the hard-thresholding operator \cite{BD}. On each component
    \begin{equation} \label{HT}
    H_{\gamma}(w_i) =
    \begin{cases}
     \  0 & {\rm if}\ |w_i| \leq \sqrt{2\gamma}\\
     \ w_i & {\rm if}\ |w_i| > \sqrt{2\gamma}.
    \end{cases}
    \end{equation}
For the $\ell_1$ case, it is also clear that
    \begin{align*}
    u^{t+1} = S_{\lambda / \beta}(w^t),
    \end{align*}
where $S_{\gamma}$ is the soft-thresholding operator \cite{DMD}
   \begin{equation} \label{ST}
    S_{\gamma}(w_i) =
    \begin{cases}
     \  w_i+\gamma & {\rm if}\ w_i \leq -\gamma\\
     \  0 & \rm{if}\ |w_i| < \gamma\\
     \  w_i-\gamma & {\rm if}\ w_i \geq \gamma.
    \end{cases}
    \end{equation}
We also consider the transformed $\ell_1$ (TL1) penalty \cite{ZX}, which nicely interpolates the $\ell_0$ and $\ell_1$ penalties:
  \begin{align*}
    \rho_a(x) = \frac{(a+1)|x|}{a+|x|}
  \end{align*}
to each component of a vector, where $a$ is a positive parameter. It is clear that 
    \begin{align*}
    \lim_{a \to 0^+}\rho_a(x) = I_{\{x \ne 0\}},   \quad \lim_{a \to +\infty}\rho_a(x) = |x|.
    \end{align*}
By solving the problem with TL1 penalty, we can also get a thresholding operator $T_{a,\gamma}$ in closed form \cite{ZX}:
    \begin{equation} \label{TT}
    T_{a,\gamma}(w_i) =
    \begin{cases}
     \  0 & {\rm if}\ |w_i| \leq t\\
     \  g_{a,\gamma}(w_i) & {\rm if}\ |w_i| > t,
    \end{cases}
    \end{equation}
where
  \begin{align*}
    g_{a,\gamma}(x) = {\rm sgn}(x) \left\{\frac{2}{3}(a+|x|)\cos\left(\frac{\phi(x)}{3}\right)-\frac{2a}{3} +\frac{|x|}{3}\right\}
  \end{align*}   
and $\phi(x)=\arccos\left(1-\frac{27\gamma a(a+1)}{2(a+|x|)^3}\right)$. Here the parameter $t$ depends on $\gamma$ as:
    
    \begin{equation}\label{threshold}
    t = \begin{cases}
     \  \gamma\frac{a+1}{a} & {\rm if}\ \gamma \leq \frac{a^2}{2(a+1)}\\
     \  \sqrt{2\gamma(a+1)}-\frac{a}{2} & {\rm if}\ \gamma > \frac{a^2}{2(a+1)}.
    \end{cases}
    \end{equation}
\medskip

\begin{remark}
It follows from the Theorem above that the limit point $(\bar{u}, \bar{w})$ satisfies the {\it equilibrium} equations for the $\ell_0$, $\ell_1$ and transformed-$\ell_1$ penalties respectively as:
\ba
\bar{u} & = & H_{\lambda /\beta}(\bar{w}),\;\;{\rm or}\;\;  S_{\lambda /\beta}(\bar{w}), \;\; {\rm or}\;\; T_{a,\lambda/\beta}(\bar{w}); \no  \\
\nabla f(\bar{w}) &=& \beta \, (\bar{w} - \bar{u}). \label{ste}
\ea
The system (\ref{ste}) 
serves as a novel   
``critical point condition''. 
This is particularly useful in the $\ell_0$ case where 
the Lagrangian function $\mathcal{L}_{\beta}(u,w)$ is discontinuous in $u$. 
\end{remark}

\section{Experimental Results}
We apply the RVSM algorithm to convolutional neural networks to see how it brings about a sparse network. In the following experiment, we consider a convolutional neural network of 3 layers and a data set of $100 \times 100$ binary images. What we care about is the percentage of the weights which are zero after training the sparse network. Many of the algorithms can result in a sparsity of over $90\%$, which means only less than $10\%$ of the parameters contribute to the model. This makes our model far more efficient than the original one without regularization.

    \begin{figure}[h!]
      \centering
      \includegraphics[width=0.8\textwidth]{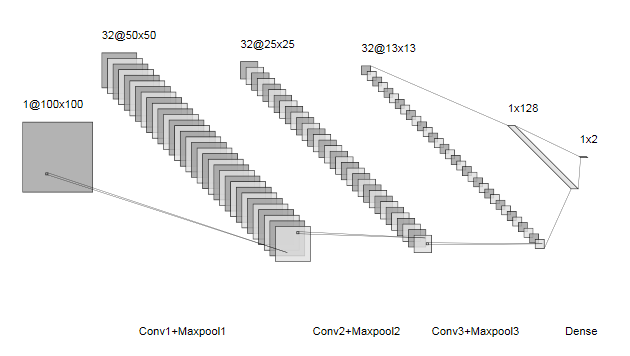}
        \caption{CNN architecture in this study.}
      \label{cnn}
    \end{figure}
    
In order to find out how the weights are distributed in each layer, we go through the structure of the network. Figure \ref{cnn} shows the number of nodes in each layer, from which we can simply calculate the number of weights needed to connect the nodes.\footnote{When generating the figure, we used a tool by Alex Lenail available at http://alexlenail.me/NN-SVG/LeNet.html}.We apply 32 $3 \times 3$ filters to the initial image to get the first convolutional layer, which results in $32 \times 3 \times 3 = 288$ weights. Similarly, each of the second and the third convolutional layer contains $32 \times 32 \times 3 \times 3 = 9216$ weights, if we apply 32 $3 \times 3$ filters again. After each convolutional layer, we add one max pooling layer with a $2 \times 2$ filter and a stride of 2. The dimension of each image is not changed after each convolution, since we have applied padding. But it is reduced by a half on both the width and the height after max pooling because of a stride of 2. Thus the dimenson of the image is reduced from $100 \times 100$ to $50 \times 50$, to $25 \times 25$ and finally to $13 \times 13$. So this produces $13 \times 13 \times 32 \times 128 = 692224$ weights when constructing a dense layer of 128 nodes. Finally, $128 \times 2=256$ weights are used to connect the dense layer to the output layer of 2 nodes, if our goal is to classify the images into two categories. From the above discussion, we notice that $97.3 \%$ of the weights are concentrated to the dense layer. We will see that most of them contribute nothing to the model after we train the sparse network. 
    
\begin{figure}[h!]
      \centering
      \includegraphics[width=0.5\textwidth]{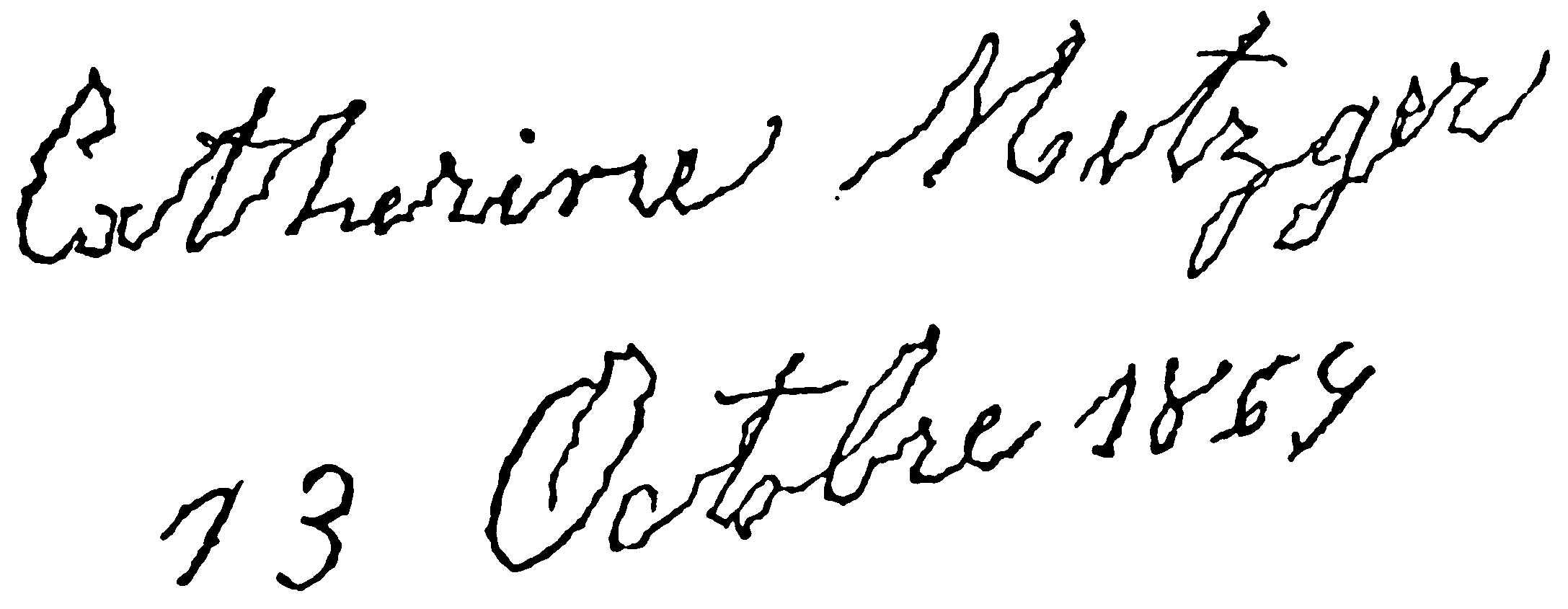}
    \caption{Handwritten sentence by a PD patient.}
      \label{PD}
    \end{figure}

The first data set we use is the images of the handwritten alphabet by Parkinson’s disease (PD) patients and normal handwritten alphabet. We know that many PD patients may suffer from tremors in their daily life and work. One remarkable feature is that the words they write can be much shakier than the normal, which can be used to distinguish a PD patient during diagnosis. Figure \ref{PD} \footnote{https://en.wikipedia.org/wiki/Micrographia\_(handwriting)} shows one real example of handwritten sentence by a PD patient.

\begin{figure}[h!]
      \centering
      \includegraphics[width=0.5\textwidth]{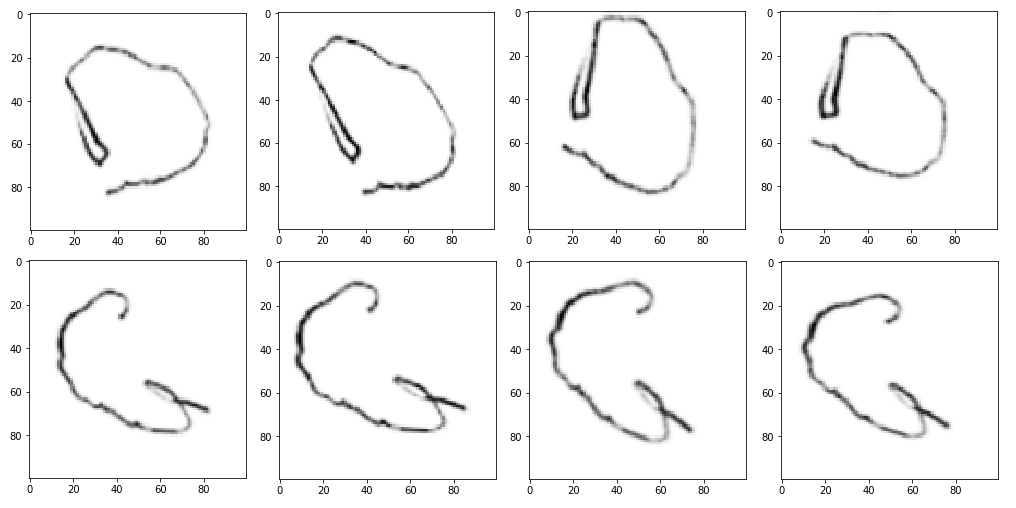}
      \caption{Sample images of PD patients' handwriting.}
      \label{letter}
    \end{figure}

From our point of view, these two writing styles -- normal vs. shaky -- can be treated as two fonts. There is one Parkinson's font available on the internet \footnote{https://www.dafont.com/parkinsons.font}, which  contains the whole alphabet of the 52 uppercase and lowercase letters. We simulate a training set of 5,000 observations and a test set of 1,000 observations by adding some rotations, affine transformations and elastic distortions \cite{Simard}. As we have mentioned, this is a data set of $100 \times 100$ binary images, of which some samples are shown in Figure \ref{letter}. Though our model is used to distinguish the letters written by a Parkinson's disease patient in this single experiment, it can be simply applied to classify any other fonts.

As most of the redundancy appears in the dense layer, we apply the threshold step of the algorithm to the weights in dense layer only. This is because if we use the same $\lambda$ and $\beta$  in all the layers, the proportion of zero weights in the convolutional layers might be high, where the zero weights can indeed grade the model. Compared to the dense layer of 700,000 weights, there is not much freedom to modify the convolutional layer of 10,000 weights. Too much sparsity leads to a sizable loss of accuracy. 

In our models, we have the freedom to set the thresholding parameters, namely $\beta$, $\lambda$ and $a$. A higher threshold usually means more sparsity, since more weights are forced to zero by the threshold. From the formula \eqref{HT} - \eqref{ST} for the $\ell_0$ and $\ell_1$ penalties, it is clear that the larger $\lambda$ is and the smaller $\beta$ is, the higher the threshold $\gamma$ will be. Given the same thresholding parameter $\gamma$, the $\ell_0$ model may result in a  sparser model than $\ell_1$, since its threshold is a square root of $\gamma$, which is higher. From the formula \eqref{TT} - \eqref{threshold} for the TL1 penalty, the smaller $a$ is, the higher the threshold is. As discussed in the previous section, when $a$ goes to infinity, TL1 becomes $\ell_1$. When $a$ goes to $0$, it becomes $\ell_0$. So as to achieve more sparsity, we may choose a small $a$.

\begin{figure}[h!]
      \centering
      \includegraphics[width=0.5\textwidth]{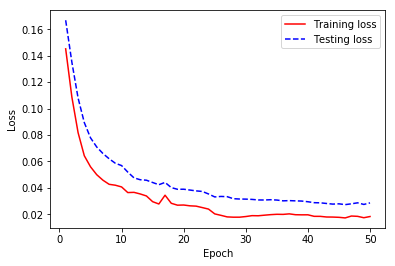}
    \caption{Training and testing loss functions vs. epochs.}
      \label{loss1}
    \end{figure}

Our algorithm converges quickly after a few iterations. In most of the cases, it obtains an accuracy of $95\%$ and a sparsity of $60\%$ after 10 epochs. The accuracy soon goes up to $98\%$ within 20 epochs, while some models achieve a sparsity of around $90\%$ eventually. Figure \ref{loss1} shows the convergence of the training algorithm.

Table \ref{res1} shows our results of sparsity and testing accuracy. It verifies what we discussed on the thresholding parameter. That is, when the threshold grows higher, the sparsity also grows correspondingly. When $a$ is less than 0.1, we achieve a sparsity of $86\%$, while the accuracy remains high. The key point should be noticed is that these sparse networks achieve almost the same, or even better accuracy than the non-sparse model. Thus we affirm that around $90\%$ of the parameters are redundant, as they hardly contribute to the accuracy of the model.

\begin{figure}
      \centering
      \includegraphics[width=0.7\textwidth]{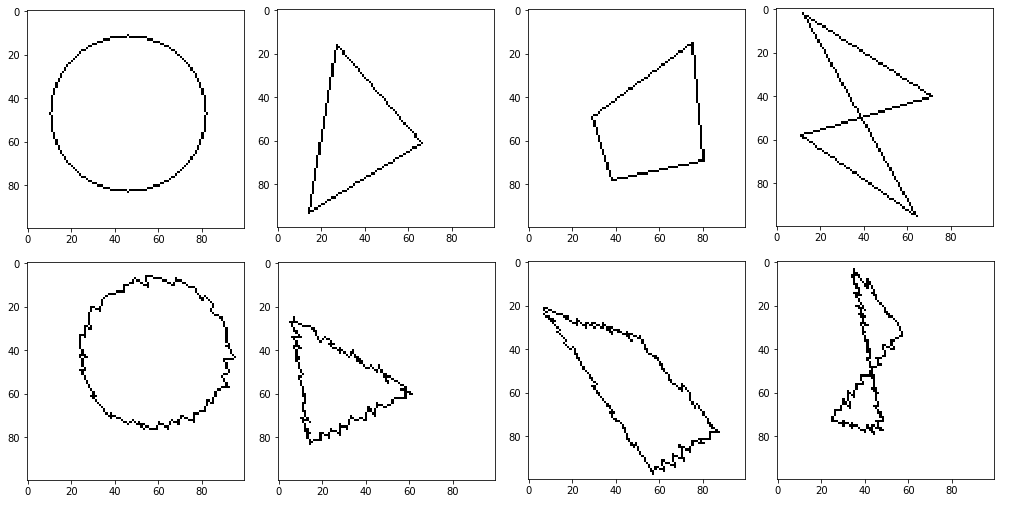}
    \caption{Normal vs. shaky shapes.}\label{shapes}

      \centering
      \includegraphics[width=0.8\textwidth]{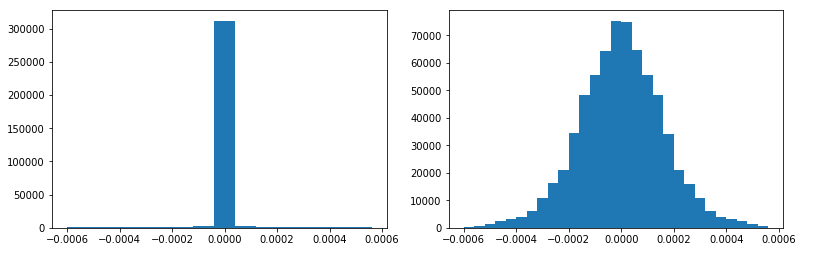}
    \caption{Distribution of weights: Sparse vs. Non-sparse networks}\label{hist1}
    \end{figure}

Another data set we consider is the images of normal vs. shaky planar shapes like triangles and quadrangles (not necessarily convex). It can be viewed as another demonstration of PD patients' handwriting, as what they draw are somehow shaky, likewise the letters they write. This data set of $100 \times 100$ binary images is simulated by adding random noise to the normal planar shapes. Figure \ref{shapes} shows some sample images of our shapes. The results on this data set are similar to those of the first data set, as shown in Table \ref{res2}. So RVSM also achieves high accuracy and sparsity on multi-scale planar curve data.

More properties of our sparse networks are as follows. First, there is a remarkable difference in distributions of the weights between the sparse and non-sparse models. For the sparse model, most of the weights are zero, while the rest are very close to zero. So its distribution looks like a vertical line plus some noise on the interval close to zero. In our example of non-sparse model, it also has a peak at zero. However, very few weights are exactly zero. Many of them are merely close to zero, while a large proportion are far away from zero. What's more, the distribution of this non-sparse model seems to be bell shaped. The distributions are shown in Figure \ref{hist1}, where the weights are normalized for better viewing.

What we also notice is that, RVSM performs much better than applying SGD directly to the TL1 penalized loss functions. As shown in Table \ref{res3}, most of the normalized weights in the SGD model are distributed between $10^{-5}$ and $10^{-3}$. It seems there is no apparent criterion to judge if a weight of $10^{-4}$ should be set to zero or it does contribute to the network. However, for the RVSM method when $a=0.01$, it is clear that $8.7\%$ of the weights are greater than $10^{-4}$ and $84.9\%$ of the weights are less than $10^{-10}$. There is a significant gap between the two scales of $10^{-4}$ and $10^{-10}$, which makes it reasonable to set all the weights less than $10^{-10}$ to zero. This leads to a network of $84.9\%$ sparsity. Another point worth mentioning is that  applying SGD directly to the penalized loss function may hurt the accuracy a lot at $a=0.01$, resulting in  $96.7\%$ accuracy for the model. This is because when $a$ is small, the penalized term behaves like $\ell_0$, which renders the objective function nearly singular. 
RVSM resolves this issue by making the penalty implicit to a thresholding process, which gives an accuracy of $99.5\%$.

Table \ref{res4} shows another interesting phenomenon. Since the weights are randomly initialized with mean zero, there is roughly even split of plus/minus signs in all layers. At the end of training, we counted the number of sign changes in the kernel of each convolutional layer, and found that  more weights changed signs in the first convolutional layer than in the next two layers. This is consistent with the network filters structured towards low pass in depth after training. 

    \begin{table}[h!]
    \caption{Testing sparsity and accuracy for the data of alphabets.} \label{res1}
    \centering
      \begin{tabular}{ m{1.5cm} m{1cm} m{1cm} m{2cm} m{2cm}  m{2cm}}
        \hline
        $\lambda$ & $\beta$ & a & Penalty & Sparsity (\%) & Accuracy (\%)\\ 
        \hline
        0.0005 & 0.1 & 0 & $\ell_0$ & 86.1 & 99.4 \\
         &  & 0.01 & TL1 & 87.6 & 99.0 \\
         &  & 0.1 & TL1 & 85.8 & 99.7 \\ 
         &  & 1 & TL1 & 78.1 & 99.3 \\ 
         &  & 100 & TL1 & 82.0 & 99.3 \\ 
         &  & $\infty$ & $\ell_1$ & 76.5 & 99.0\\
        \hline
      \end{tabular}
    \end{table}
    
    \begin{table}
    \caption{Testing sparsity and accuracy for the data of planar shapes.}
    \centering
      \begin{tabular}{ m{1.5cm} m{1cm} m{1cm} m{2cm} m{2cm}  m{2cm}}
        \hline
        $\lambda$ & $\beta$ & a & Penalty & Sparsity (\%) & Accuracy (\%)\\ 
        \hline
        0.0005 & 0.1 & 0 & $\ell_0$ & 90.2 & 99.9   \\
         &  & 0.01 & TL1 & 83.5 & 99.1 \\
         &  & 0.1 & TL1 & 87.6 & 99.8 \\ 
         &  & 1 & TL1 & 74.9 & 99.9 \\ 
         &  & 100 & TL1 & 75.0 & 99.9 \\ 
         &  & $\infty$ & $\ell_1$ & 74.6 & 99.6 \\ 
        \hline
      \end{tabular}
      \label{res2}

    \end{table}
    \begin{table}
    
    \caption{Sparsity and accuracy: RVSM vs. Direct SGD for TL1 penalty}
    \centering
      \begin{tabular}{ m{1cm} m{2cm} | m{1cm} m{1cm} m{1cm}  m{1cm}  m{1cm} m{2cm}}
        \hline
         &  & \multicolumn{5}{c}{Sparsity (\%) of $10^{-n}$ scale}  \\
         a & Algorithm & $10^{-2}$ & $10^{-3}$ & $10^{-4}$ & $10^{-5}$ & $10^{-10}$ & Accuracy (\%)\\ 
        \hline
         0.01 & RVSM & 99.7 & 96.0 & 91.3 & 88.6 & 84.9 & 99.5\\
         0.01 & SGD & 99.9 & 99.9 & 45.9 & 5.44 & $10^{-5}$ & 96.7\\ 
        \hline
         100 & RVSM & 99.9 & 97.5 & 92.7 & 88.5 & 80.3 & 99.3\\
         100 & SGD & 99.9 & 99.7 & 48.1 & 6.68 & $10^{-5}$ & 99.0\\ 
        \hline
      \end{tabular}
      \label{res3}
          
          \vspace{1cm}
          \caption{Number of sign changes and relative \% in kernels of convolutional layers.}
    \centering
      \begin{tabular}{ m{2cm} m{2cm} m{2cm}  m{2cm}}
        \hline
        a & layer 1 & layer 2 & layer 3 \\ 
        \hline
        0.01 & 72 (25.0\%)  & 1120 (12.2\%)  & 769 (8.34\%)\\
        1 & 45 (15.6\%)  & 1133 (12.3\%)  & 784 (8.51\%)\\
        100 & 35 (12.2\%) & 1001 (10.9\%) & 995 (10.8 \%)\\
        \hline
      \end{tabular}
      \label{res4}
    \end{table}

\section{Conclusions}

In this paper, we have applied the RVSM algorithm to learn sparse neural networks. We have achieved an accuracy of $99\%$ and a sparsity of $87\%$ when training CNNs on a data set consisted of synthetic handwritten letters and planar curves by PD patients, and normal handwriting. We have also discussed the tuning of thresholding parameters, and verified the fact that a higher threshold can produce higher sparsity. What's more, our experiments show that the RVSM outperforms the direct application of SGD on the penalized loss function, in both sparsity and accuracy. The RVSM generates a significant gap between the weights of large scale and small scale, which acts as an indicator to show sparsity. 

\subsubsection{Acknowledgements.}
The work was partially supported by NSF grant IIS-1632935. 
The authors would like to thank Profs. Xiang Gao and Wenrui Hao at Penn State Universty for helpful discussions of handwritings and drawings on neuropsychological exams and diagnosis. 

%\bibliographystyle{plain}
%\bibliography{main}

\end{document}